# Unsupervised classification to improve the quality of a bird song recording dataset

Félix Michaud [1,*], Jérôme Sueur [1], Maxime Le Cesne [1], Sylvain Haupert [1]

[1] Institut Systématique, Évolution, Biodiversité (ISYEB), Muséum national d'Histoire naturelle, CNRS, Sorbonne Université, EPHE, Université des Antilles, 57 rue Cuvier, 75005 Paris, France

* corresponding author Institut Systématique, Évolution, Biodiversité (ISYEB), Muséum national d'Histoire naturelle, CNRS, Sorbonne Université, EPHE, Université des Antilles, 57 rue Cuvier, 75005 Paris, France, +33 1 40 79 33 98, felix.michaud@mnhn.fr

**Highlights**

- Open audio databases provide unique information for ecological sciences
- However, most of them are weakly labelled with a risk of signal confusion
- A labelling function based on segmentation, featurization, and clustering is proposed
- A test on the Xeno-Canto database revealed a significant reduction of label error
- Unsupervised labelling can foster the use of audio archives

**Abstract**

Open audio databases such as Xeno-Canto are widely used to build datasets to explore bird song repertoire or to train models for automatic bird sound classification by deep learning algorithms. However, such databases suffer from the fact that bird sounds are weakly labelled: a species name is attributed to each audio recording without timestamps that provide the temporal localization of the bird song of interest. Manual annotations can solve this issue, but they are time consuming, expert-dependent, and cannot run on large datasets. Another solution consists in using a labelling function that automatically segments audio recordings before assigning a label to each segmented audio sample. Although labelling functions were introduced to expedite strong label assignment, their classification performance remains mostly unknown. To address this issue and reduce label noise (wrong label assignment) in large bird song datasets, we introduce a data-centric novel labelling function composed of three successive steps: 1) time-frequency sound unit segmentation, 2) feature computation for each sound unit, and 3) classification of each sound unit as bird song or noise with either an unsupervised DBSCAN algorithm or the supervised BirdNET neural network. The labelling function was optimized, validated, and tested on the songs of 44 West-Palearctic common bird species. We first showed that the segmentation of bird songs alone aggregated from 10% to 83% of label noise depending on the species. We also demonstrated that our labelling function was able to significantly reduce the initial label noise present in the dataset by up to a factor of three. Finally, we discuss different opportunities to design suitable labelling functions to build high-quality animal vocalizations with minimum expert annotation effort.

# 1. Introduction

Bird song has been the center of most bioacoustic studies for about 70 years and is currently a key element of soundscape ecology and ecoacoustic research in terrestrial environments (Xie et al., 2020). The sounds produced by birds, essentially their territorial songs, are commonly recorded using parabolic or directional microphones so that a single individual representing a single species appears as the sound of highest intensity. Such recordings, known as focal recordings, are deposited and shared in public and private sound libraries whose sizes are increasing continuously (Toledo et al., 2015).

Xeno-Canto is the most used and diversified focal bird song database (Xeno-canto, 2021). This community project aims at gathering and curating bird sounds from expert and non-expert contributors with a total of around 678,000 audio recordings of 10,300 species from all major world areas. Xeno-Canto can be used for taxonomy, behavior, biogeography and ecology research (Planqué and Vellinga, 2008). In recent years, Xeno-Canto has been the major source of machine and deep learning challenges for the automatic identification of bird species through acoustics (Joly et al., 2015). Algorithms to monitor bird species (Denton et al., 2022; Dufourq et al., 2022; Nolasco et al., 2022; Stowell, 2022), study dialects (Cohen et al., 2022), or identify individuals (Martin et al., 2022) are also growing quickly. Among them, the convolutional neural network BirdNET currently appears as the most promising open-source and user-friendly solution for automatic bird identification (Kahl et al., 2021).

Xeno-Canto is extremely useful but is weakly labelled: a species name is attributed to each recording without timestamps to provide temporal localization of bird song sequences. Three main methods have been used to reduce this inaccuracy by temporally localizing the sounds of interest. The first option consists in asking bird experts to manually annotate the recordings. By listening to and visualizing spectrographic representation of the audio, the experts select the beginning and the end of each sound of interest, sometimes manually drawing bounding boxes on the spectrogram. This procedure ensures high quality data but requires rare expertise, is time-consuming, and can be complexified by possible inconsistencies between labels added by different experts (Donmez and Carbonell, 2008; Gil-González et al., 2021; Sheng et al., 2008). The second method consists in semi-automatic annotation, combining automatic clustering and manual annotation. This approach has already been applied for a wide range of animal vocalizations, including soundscape analysis (Arthur et al., 2021; Fukuzawa et al., 2020; Phillips et al., 2018; Ruff et al., 2021; Steinfath et al., 2021; Zhong et al., 2021). In this work, we focus on a third method that requires no direct involvement of human experts but instead develops a program that automatically assigns a label to each sound unit. This type of program is designated as a labelling function (Ratner et al., 2016; Zhang et al., 2022).

In the context of building a custom bird song dataset from open databases, most of the labelling functions consist of automatically localizing the salient sounds by applying segmentation methods, from simple energy threshold (Fodor, 2013; Lasseck, 2013; Potamitis, 2014) to more complex signal and image processing (Sprengel et al., 2016). Then, the weak label of the identified bird is assigned automatically to all the segmented sounds. However, other sounds than the one of interest might be segmented. Such wrong label assignment, known as label noise, has been pointed out as one of the main limitations in the process of developing a robust bird sound detection algorithm (Kahl et al., 2021). Indeed, the presence of label noise in datasets can impair classification tasks (Frenay and Verleysen, 2014) by degrading the learning performance of the algorithm or even preventing the convergence of the model (Arpit et al., 2017; Chen et al., 2019; Karimi et al., 2020; Sheng et al., 2008), and can also induce class imbalance problems.

Three solutions were proposed to reduce label noise (Frenay and Verleysen, 2014). The first solution relies on the implicit capacity of certain algorithms to be robust to noise during the training session. As an example, it was shown that very large artificial neural networks are able to fit a dataset with label noise up to 90% (Rolnick et al., 2017). However, this model-centric option requires a large amount of data and computer power, lacks easy deployment capacity, and increases computing time, energy consumption, and $CO_2$ emissions (Schwartz et al., 2020; Strubell et al., 2019). This solution was largely chosen for bird sound detection and classification (Joly et al., 2021; Stowell, 2022). The second solution consists of the explicit handling of label noise by an algorithm (Algan and Ulusoy, 2021; Frenay and Verleysen, 2014). Such a solution is denoted as a noise-tolerant or noise model-based method and requires a good prior knowledge of the label noise to integrate a representation of the noise in the model. Here, we focus on a third solution that aims at addressing label noise at the source with an upstream filter that detects and removes mislabeled data from the dataset (Brodley et al., 1999). This way, by directly removing label noise at the source, the learning process can be facilitated (Sambasivan et al., 2021), which is in line with the data-centric approach that advocates for cleaner and smaller datasets to create good conditions for lighter training processes (Motamedi et al., 2021; Ng et al., 2021).

To address this issue and reduce label noise in large bird song datasets, we introduce a data-centric novel labelling function composed of three successive steps: 1) time-frequency sound unit segmentation, 2) feature computation for each sound unit, and 3) classification of each sound unit as bird song or noise with either an unsupervised DBSCAN algorithm or the supervised BirdNET neural network. The labelling function was optimised, validated, and tested on the songs of 44 West-Palearctic common bird species.

## 2. Material and methods

An overview of the complete workflow to design the labelling function is presented Figure 1.

*2.1. Audio database*

The audio database included West-Palearctic common bird species of the Passeriformes, Columbiformes, Piciformes, Cuculiformes, and Strigiformes orders. The monitoring of these species is crucial for biodiversity assessment as testified by the Pan-European Common Bird Monitoring Scheme (PECBMS, 2021). The audio database was split into three sets: a training set, a validation set, and a test set. The training set and the validation set referred to the same 22 bird species (Table A.1.). The selection of the species followed a previous study conducted in the Parc Naturel Régional of Vallée de Chevreuse (24,300 ha), a protected area located 40 km southwest of Paris, France (Depraetere et al., 2012). The third set included 22 other common species that were selected from the Global Biodiversity Information Facility (GBIF, 2021) database (Table A.2.).

All recordings were automatically retrieved from Xeno-Canto. This international online archive gathers focal bird song recordings. To obtain a representative dataset of bird sounds within a reasonable annotation time, the first step was to retrieve a maximum of 100 files for each species with the following metadata conditions: (1) song type, (2) high quality rate (A or B), and (3) a duration between 20 s and 180 s. Twenty files were then randomly chosen for each species. Finally, 20 s of each file were selected at the start of the first song encountered in the file and annotated.

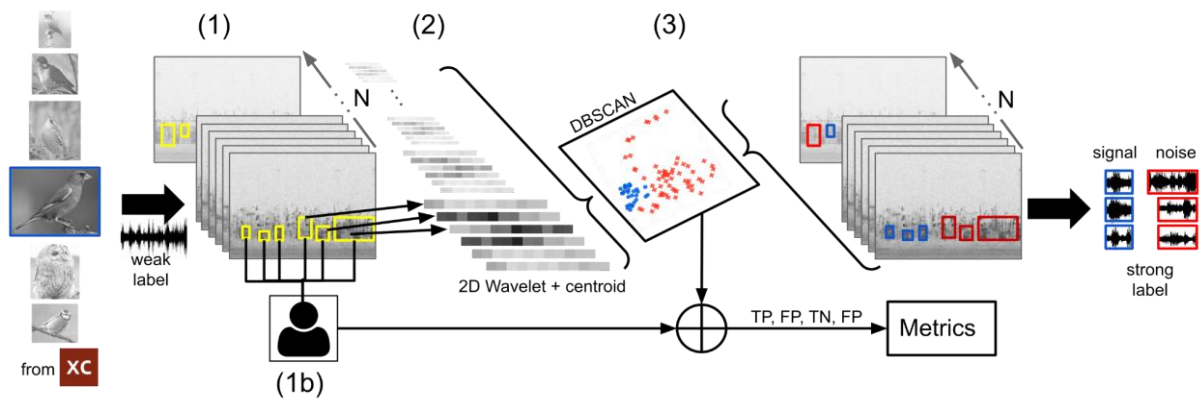

Figure 1. From weak to strong labels: workflow of the labelling function. After collecting N weakly labelled audio recordings of a focal species from the Xeno-Canto database, the labelling function workflow consists of (1) segmenting each audio recording into time-frequency sound units, (2) calculating 49 features (2D wavelet coefficients (n = 48) + frequency centroid (n = 1)) of each sound unit, (3) clustering all sound units into signal

(bird song belonging to the focal species) or noise (everything else) with DBSCAN. Finally, a strong label is attributed to each sound unit. In parallel, an expert annotates all sound units (1b) to calculate the metric to evaluate the performance of the labelling function. TP stands for True Positive, FP for False Positive, TN for True Negative, FP for False Positive

*2.2. Automatic segmentation of the regions of interest (ROIs)*

2.2.2. Pre-processing

The mp3 audio files were loaded with a sampling rate of 44.1 kHz. A band-pass filter from 100 Hz to 18,000 Hz was applied to remove low-frequency background noise as well as high-frequency artefacts due to mp3 compression. The short-term fourier transform (STFT) was calculated with a Fourier window of 2,048 samples, an overlap of 1,024 samples and a Hanning window (Fig. 2a). The resulting spectrogram was then resized in time and frequency so that closely related pixels were merged together (Fig. 2b). Background noise was partly removed by estimating the stationary noise directly on the spectrogram. For each frequency band, the average amplitude along the time axis was estimated to obtain the raw noise profile along the frequency axis. Then, the noise profile was smoothed using a running mean over 25 frequency bands. Eventually, the noise profile was subtracted to the overall resized spectrogram (Towsey, 2013) and the amplitude values of the resulting spectrogram were finally turned into dB (Fig. 2c).

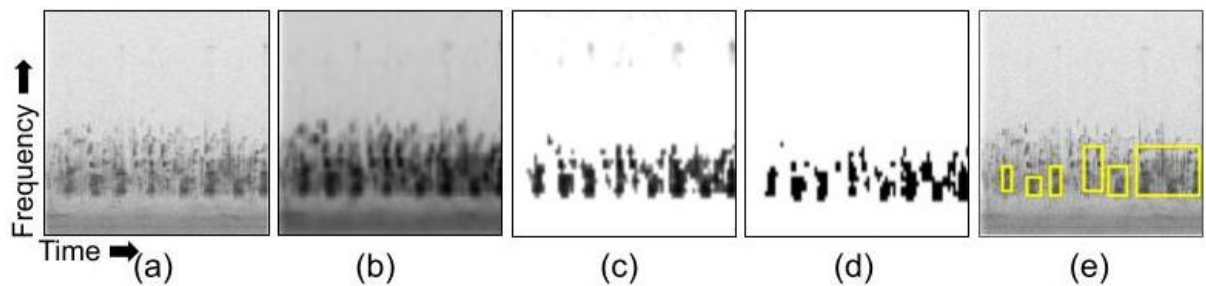

Figure 2. Spectrogram pre-processing and segmentation. Audio files were first converted into spectrograms (a), spectrograms were then resized in time and frequency (b). After that, noise profiles were subtracted from spectrograms and the values were turned into dB (c). Furthermore, spectrograms were binarized with the hysteresis thresholding binarization method (d). The last step is the delimitation of the bounding boxes (e).

2.2.3. Segmentation

Segmentation was processed on the spectrogram by delimiting time-frequency rectangular bounding boxes around salient sounds. The delimitation of these regions of interest (ROIs) was based on a region growing segmentation method known as hysteresis thresholding binarization (Ulloa et al., 2021). The method is based on two dB thresholds: the high threshold selects high intensity pixels while the low threshold groups neighbor pixels (Fig. 2d). The bounding boxes that were separated in time by less than 0.24 s and separated in frequency by less than 170 Hz, were merged. The bounding boxes that lasted less than 0.36 s were not considered as they are too short to be a bird song unit (Fig. 2e).

2.2.4. Optimization and metrics

The pre-processing and segmentation processes depended on four parameters: the time and frequency scaling factor of the spectrogram and the two thresholds of the hysteresis binarization. The optimal combination of these four parameters was estimated on a selection of files. Manual delimitation of bounding boxes was conducted on these files using the Audacity time-frequency annotation tool on the spectrogram view (Audacity v3.1.3, 2021). Then, the automatic extraction workflow was applied for each species on the specific annotated 20 s for a list of combinations of the four parameters. The resizing factor in time ranged from 6 to 18, the resizing factor in frequency ranged from 4 to 30, the upper threshold ranged from 10 to 40 dB, and the lower threshold ranged from 7 to 37 dB.

The distance between manual and automatic delimitations was measured with the Intersection Over Union (IOU), a metric that measures area overlap between two boxes (Zou et al., 2019). The median of the IOUs was calculated for each parameter combination and each file and then for all the files for a species. The best IOU score for all species was then selected to obtain the best combination of parameters.

Once the best combination for the four parameters was found, all the ROIs present in the 20 s audio segments for the 20 files for all species were automatically extracted based on the bounding boxes' limits. The ROIs were delineated in time by the vertical limits and in frequency by the horizontal limits. The sound was then extracted first by trimming the audio recording around the beginning and end of the ROI in time and then by applying a fifteenth order bandpass butterworth filter with the lower and higher cutoff frequencies corresponding to the lower and upper spectral limits of the ROIs.

*2.3. Automatic classification of the regions of interest*

2.3.1. Features

The unsupervised classification was performed on the spectrogram. However, the spectrogram resolution and the bandpass filter differ from those used for the ROI segmentation. The sample rate was set to 24 kHz, the Fourier window took 512 samples, and the bandpass filter range was between 250 Hz and 11 kHz. The unsupervised classification was based on two signal descriptors. The first descriptor was a vector of values resulting from a two-dimensional wavelet analysis (Ulloa et al., 2018). The regions of interest were convolved with a series of scale Gabor filters to extract spectro-temporal characteristics according to the selected resolution. It was composed of 48 Gabor filters with two vertical filters, two horizontal filters, and four diagonal filters with two angles at 45° and 135° at six different scales. The second descriptor was based on a Fourier analysis that consisted in measuring the spectral centroid of each region of interest. The feature vector to describe a ROI therefore had dimensions of 49 by 1. A single vector is obtained for every ROI.

2.3.2. Clustering

The classification of bird songs and noise was achieved with the unsupervised method called Density-based spatial clustering of applications with noise (DBSCAN) (Ester et al., 1996). We opted for an unsupervised method for the workflow to be applied on any bird species with enough audio recordings without the need for a learning stage. Referring to a density-based definition of clusters, DBSCAN does not require an initial number of clusters and works for non-spherical clusters. In addition, DBSCAN allows the identification of outliers that can be considered as noise, a prerequisite of labelling functions. DBSCAN searches for core samples by starting with an arbitrary point *g* and retrieves all points that are distant by less than an Euclidean distance (*Eps*) and have a minimum number of points in their neighborhood (*MinPts*). Samples are defined as noise if they are outside any cluster. There are no ideal *Eps* and *MinPts* values so their setting is not trivial (Schubert et al., 2017). *Eps* was estimated individually for each species using a K-dist graph, which plots the average distance between each point and its *k* nearest neighbors in an ascending order. The inflection point, or "knee", indicates the limit between the maximum distance of the core points of the cluster (signal) and the outlier points of the cluster (noise). This point was automatically detected with a knee locator algorithm called *Kneed* (Satopaa et al., 2011).

*MinPts* was determined by running the clustering algorithm of each species on the training set with *MinPts* values ranging from 1% to 50% of the number of ROIs. The macro average precision over all species was calculated for each percentage. This led to *MinPts* of 10% for all sets.

To evaluate the result of the classification, the points belonging to the cluster were annotated as positive samples and the outliers as negative samples. In cases where there were more than one cluster, we kept only the largest cluster as we formed the hypothesis that the other smaller clusters were bird songs due to other species, anthropogenic noise such as car traffic, or unusual bird songs that could belong to the species of interest. Knowing the false and true positives and negatives, we applied the three performance metrics commonly used in machine learning to evaluate the automatic process: (1) macro-average of precision (number of true positives over the sum of true positives and true negatives), (2) macro-average of recall (number of true positives over the sum of true positives and false negatives), and (3) macro-average of F1 score (combination of the precision and recall). Finally, we introduced two other metrics. The initial percentage of noise corresponds to the number of ROIs labelled as noise over the total number of ROIs. The final percentage of noise corresponds to the number of false positives over the sum of the true and false positives, equivalent to 1-precision.

2.3.3. Manual annotation

The 5,315 regions of interest returned by the automatic segmentation were manually annotated by one of us (MLC) as 'signal' (part or complete bird song) or 'noise' (not a bird song or a bird song overlapping with other sounds or a portion of a bird song that is too small to be recognizable by a human expert) using the Audacity spectrographic view.

*2.4. Supervised classification with BirdNET*

To compare the DBSCAN output with a supervised method, the classification of bird song was operated with the convolutional neural network (CNN) BirdNET-Lite (Kahl et al., 2021). BirdNET is designed for bird sound classification of 984 bird species around the world and is considered state-of-the-art bird audio identification. BirdNET was geographically conditioned to France during the spring season (fifteenth week of the Julian calendar). We kept the default parameters which were a sensitivity of 1 and a minimum of confidence of 0.1. The ROIs of different lengths were padded with zeros to have the same length. Then the ROIs were passed through the CNN for classification.

*2.5. Implementation*

Both methods could be run on a personal computer or on a virtual machine. In the latter case, with a single core with a clock of 2.2 GHz and 13 GB of RAM, the computation time to process the validation dataset (20 audio files of 22 species) took 7 minutes to download the files from xeno-canto, 6 minutes to extract the ROIs, 13 minutes for the ROIs classification with DBSCAN (feature extraction and clustering), and 25 minutes for the ROIs classification with BirdNET.

The complete process was written in Python language with the packages scikit-maad v1.3.0 (Ulloa et al., 2021), scikit-learn v0.23.2 (Pedregosa et al., 2011), librosa v0.8.0 (McFee et al., 2015), and kneed v0.8.1 (Satopaa et al., 2011). The full code is available as a Python package on Github (https://github.com/ear-team/bam).

## 3. Results

### *3.1. Audio files*

A total of 880 audio files of 20 s (20 audio files * 44 species) were retrieved per species from Xeno-Canto, resulting in 4h53'20" of recordings.

### *3.2. Regions of interest (ROIs)*

The best combination of the four parameters for the automatic segmentation of the ROIs was 10 and 15 for the time and the frequency downscaling factor, respectively, of the spectrogram, and 37 dB and 33 dB for the first and the second threshold, respectively, of the hysteresis binarization.

For the training set, a total of 1,855 bounding boxes were automatically extracted. The average number of ROIs per species was 84 with a minimum of 41 for the tawny owl (*Strix aluco*) and a maximum of 134 for the mistle thrush (*Turdus viscivorus*). The median percentage of noise per species estimated by the labelling function was 36.5% with a great disparity between species (Table 1, Table A.1.). The song thrush (*Turdus philomelos*) was the species with the highest initial quantity of noise (78%) and the tawny owl (*Strix aluco*) had the lowest (15%).

The validation set contained 1,732 ROIs extracted from the same 22 species as the training set. The European green woodpecker (*Picus viridis*) had the lowest number of ROIs (40) and the common cuckoo (*Cuculus canorus*) had the highest (162). The median percentage of noise per species was 35.5% (Table 1, Table A.1.). The song thrush (*Turdus philomelos*) had the largest initial quantity of noise (83%) when the European greenfinch (*Chloris chloris*) had the lowest (13%).

The test set, which was composed of 22 other species, contained 1,728 ROIs. The Cetti's warbler (*Cettia cetti*) had the lowest number of ROIs (48) and the Eurasian blue tit (*Cyanistes caeruleus*) had the largest (123). The median percentage of noise per species set was 48% (Table 1, Table A.2.). The Eurasian bullfinch (*Pyrrhula pyrrhula*) had the largest initial quantity of noise (82%) while the black redstart (*Phoenicurus ochruros*) had the lowest (10%).

### *3.3. Clustering*

In the case of the training set, a reduction in the median noise was observed for all species (Fig. 3a), from 36.5% to 12.5%. A similar reduction was observed for the validation set, from 35.5% to 13.5% (Fig. 3b, Table 1). The dispersion of the final noise was higher for the train set than for the validation set as indicated by a broader interquartile range after denoising.

|  |  | Training | Validation | Test |
|---|---|---|---|---|
|  | Median Initial noise | 36.5% | 35.5% | 48.0% |
| DBSCAN Clustering | Median Final noise | 12.5% | 13.5% | 33.5% |
|  | Macro precision | 0.81 | 0.82 | 0.67 |
|  | Macro recall | 0.65 | 0.72 | 0.68 |
|  | Macro F1 score | 0.67 | 0.73 | 0.65 |
| BirdNET | Median Final noise | 14.0% | 14.5% | 23.0% |
|  | Macro precision | 0.81 | 0.81 | 0.75 |
|  | Macro recall | 0.82 | 0.83 | 0.74 |
|  | Macro F1 score | 0.81 | 0.81 | 0.71 |

Table 1. Global scores per labelling function for the three datasets. Noise levels are expressed in percentage when quality metrics vary between 0 and 1.

After applying the DBSCAN clustering, the training set contained two species (common cuckoo [*C. canorus*] and tawny owl [*S. aluco*]) that were completely denoised and the validation set contained three species (common cuckoo [*C. canorus*], tawny owl [*S. aluco*], and long-tailed tit [*Aegithalos caudatus*]) that were fully denoised as well. One species, the song thrush (*T. philomelos*), remained with a high final noise with values between 76% in the training dataset and 77% in the validation dataset. This species was also the species with the highest initial quantity of noise (78% in the training set and 83% in the validation set).

In the case of the test set, which was composed of 22 bird species that were not in the training and validation sets, a reduction of the noise was also observed, from 48% to 33.5% (Fig. 3c, Table 1). The slighter reduction of the noise compared with the training and validation sets was associated with a wider dispersion of the final noise. After the DBSCAN clustering, only the noise included in the recordings of the meadow pipit (*Anthus pratensis*) could be reduced to zero. An increase in noise was observed for the little owl (*Athene noctua*) from 58% to 75% and for the goldcrest (*Regulus regulus*) from 44% to 53%. The Eurasian blue tit (*C. caeruleus*), the common reed bunting (*Emberiza schoeniclus*), and the Eurasian bullfinch (*Pyrrhula pyrrhula*) showed high final noise scores of 79%, 64%, and 70%, respectively.

The macro precision, which estimated the degree of cleanness of the dataset, was above 0.81 for the training dataset and above 0.82 for the validation dataset, while it dropped for the test dataset to 0.67 (Table 1). The macro recall was below 0.8 for all the three datasets, with values ranging between 0.65 for the training set and 0.72 for the validation set (Table 1). The macro F1 score had values between 0.65 for the test set and 0.73 for the validation set.

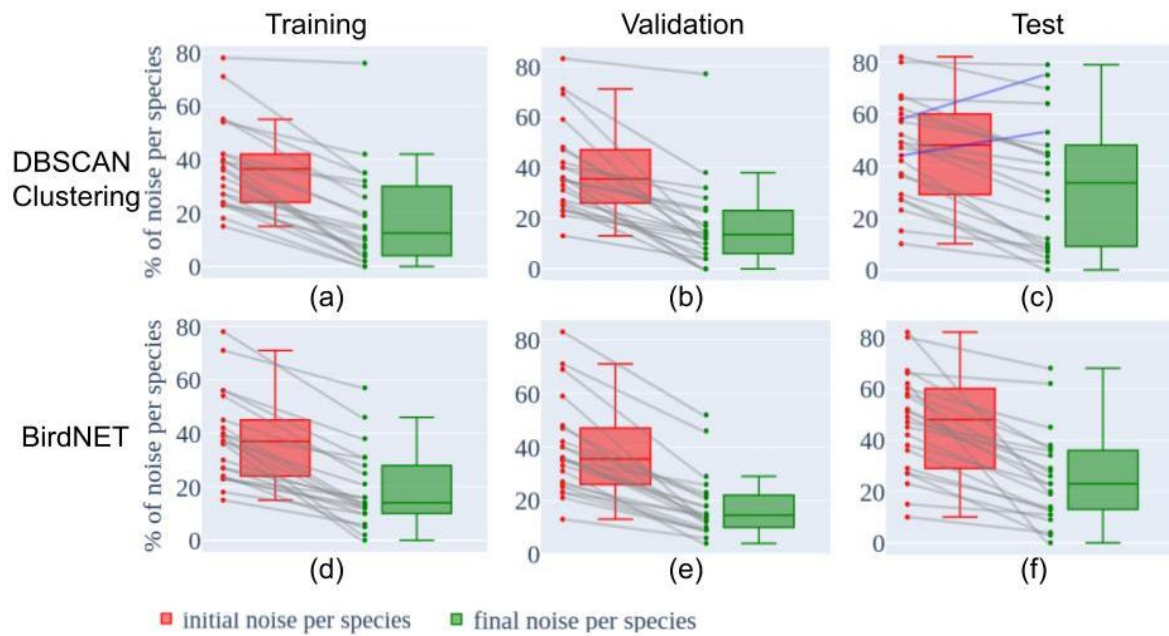

Figure 3. Initial and final quantity of noise for the three datasets when applying the DBSCAN unsupervised clustering (a, b, c) or the supervised BirdNET classification (d, e, f). The initial percentage of noise corresponds to the number of ROIs labelled as noise over the total number of ROIs. The final percentage of noise corresponds to the number of false positives (i.e., the ROIs that were automatically assigned as signal but were noise) over the sum of the true and false positives. It can also be expressed as 100% – precision (i.e., the number of true positives over the sum of true and false positives). Each box represents the median (central line) and the 25% (Q1) and 75% quartiles (Q3) (box limits), while the whiskers represent the last sample point below and above 1.5 times the interquartile range (Q3–Q1) for the lower and upper whiskers, respectively. For an exact representation of the distribution, each recording is shown with a point and lines connecting the treatments.

To understand the influence of the initial quantity of noise on the clustering quality, the initial quantity of noise was represented against the final quantity of noise (Fig. 4a). Species that had more than 40% of initial noise had a low reduction of noise, especially for the clustering labelling function. For the species with less than 40% initial noise, DBSCAN clustering was more efficient than BirdNET as more points were close to 0% final noise. All species that reached 0% noise after clustering had an initial quantity of noise under 40%.

The polynomial regression applied to the clustering results (Fig. 4a) revealed that for the species with an initial quantity of noise between 10% and 35%, the denoising completed by the clustering algorithm was particularly efficient as all these species had a final quantity of noise around 5%. Beyond 35%, the denoising was less effective as shown by the polynomial approximation which was closer to the y = x line.

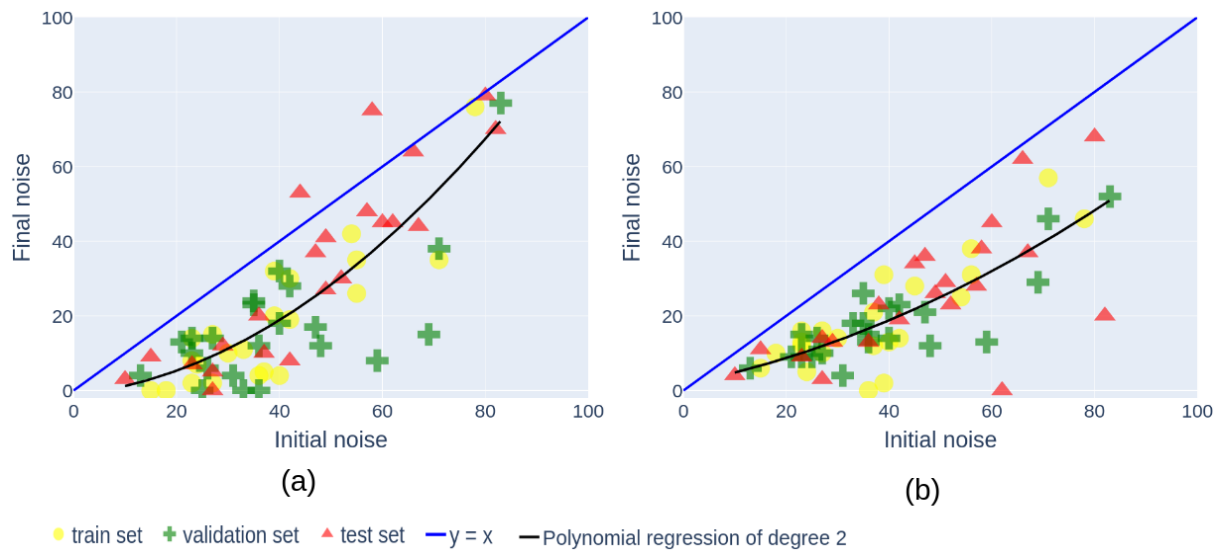

Figure 4. Comparison of initial noise with final noise with the unsupervised DBSCAN clustering method (a) and with the BirdNET algorithm (b). Each point represents one species. The y = x blue line indicates when the final quantity of noise is equal to the initial quantity of noise. The black line is a polynomial regression of degree 2 indicating the general trend.

*3.4. BirdNET*

The median final noise values obtained with BirdNET were higher than for the DBSCAN clustering algorithm with values of 14% and 14.5% for the training and validation sets, respectively, while the median final noise was lower with a value of 23% for the test set (Fig. 3d–f, Table 1). Only two species were completely denoised in the three sets [European green woodpecker (*P. viridis*] and common starling [*Sturnus vulgaris*]). Contrary to DBSCAN clustering, BirdNET classification efficiency is expected to be independent of the initial quantity of noise per species as confirmed by the quasi-linear relationship between the initial and final noise (Fig. 4b).

The macro precision was above 0.8 for the training and validation sets, which was lower than for the DBSCAN clustering. The macro precision was 0.75 for the test, which was higher than for the DBSCAN clustering (Table 1). The macro recall values ranged between 0.74 for the test set and 0.84 for the validation set, which was higher than for the DBSCAN clustering. BirdNET had a higher recall than DBSCAN clustering.

*3.5. Clustering and BirdNET*

The final noises obtained with both methods were compared for all the sets (Fig. 5). The performance of DBSCAN clustering was higher when the initial noise was below 20%. Both methods behaved in the same way when the initial noise was between 20% and 40%. The BirdNET labelling function obtained better results when being applied on species with greater than 40% initial noise.

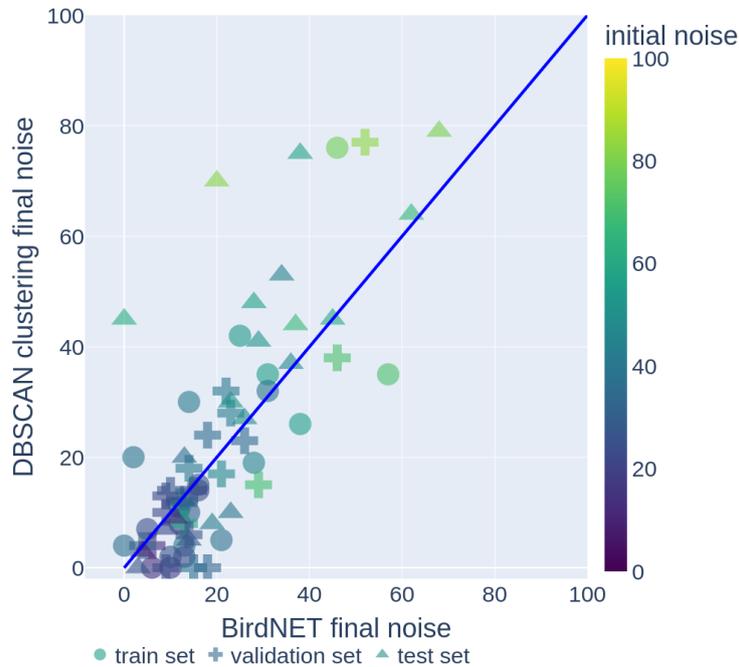

Figure 5. Diminution of noise compared to the DBSCAN clustering and BirdNET algorithms in regard to the initial quantity of noise. DBSCAN clustering was more performant for species lying below the y = x blue line. BirdNET was more performant for species lying above the y = x blue line.

## 4. Discussion

In this work, we designed a specific labelling function to improve the quality of the Xeno-Canto bird song dataset by minimizing label noise. This labelling function followed a three-step process: 1) spectrographic time-frequency segmentation, 2) feature computation, and 3) noise or signal classification with either an unsupervised (DBSCAN) or a supervised (BirdNET) algorithm. We will comment on the performance of the new labelling function, discuss the advantages and drawbacks of each step of the workflow, and finally propose possible improvements.

*4.1. Performance of the labelling function*

Through manual annotation of focal bird recordings, we could assess label noise generated by automated labelling for the first time. The proportion of segmented sounds that do not correspond to the species of interest varied between 10% as observed for the black redstart (*P. ochruros*) and 80% such as for the song thrush (*T. philomelos*), the Eurasian bullfinch (*P. pyrrhula*), and the Eurasian blue tit (*C. caeruleus*). Over all sets, 16 of the 44 species were affected by more than 50% label noise. Such a level of label noise can be explained by 1) the inherent variability induced by different audio equipment ranging from hyper focal to omnidirectional microphones, 2) different background noise levels due to other sound sources including other bird species, or 3) the diversity of the time-frequency bird song structure (Kershenbaum et al., 2016).

The labelling function could reduce the median label noise by up to a factor three with a large disparity between the species and the dataset. When considering all datasets, 25 of 44 species had a final label noise below 25% using the unsupervised method (DBSCAN), while this proportion slightly increased to 31 of 44 species using the supervised method (BirdNET). The performance of the labelling function was significantly lower for the test set than for the training and validation sets regardless of the classifying method. The difference in performance might be explained by the segmentation that generated a higher initial label noise level in the test set than in the training and validation sets. As the segmentation was optimized on the bird songs of the species in the training set, the hyperparameters found after tuning might be too specific and, hence, overfit the process.

The DBSCAN clustering success varied depending on the initial label noise. In particular, DBSCAN outperformed BirdNET when the initial label noise was below 20% but underperformed when it was above 40%. Contrary to BirdNET, the performance of DBSCAN is dependent on the proportion of label noise in the dataset. If the initial label noise level was above a critical value, the adaptive selection of *MinPts* and *Eps* might be biased by the presence of a critical amount of sounds that is not representative of the sound of interest (Latifi-Pakdehi and Daneshpour, 2021). A large initial label noise level might also increase the chance to aggregate similar audio samples into a cluster that do not belong to the species of interest.

Signal segmentation and classification of bird songs is challenging due to the high temporal and frequency diversity of acoustic units (Kershenbaum et al., 2016; Priyadarshani et al., 2018). Here, the level of label noise was particularly important for high frequency bird songs such as those of the blue tit (*C. caeruleus*) or the kinglet (*Regulus regulus*). In that case, the signal-to-noise ratio, due to strong attenuation was expected to be lower and so the regions of interest were less salient and more difficult to delimit. The results were also low for species emitting brief sounds like the bullfinch (*P. pyrrhula*) or long sequences of notes like the song thrush (*T. philomelos*). Acoustic reverberation and attenuation due to outdoor propagation can also blur the temporal architecture of the song, making the detection of fast events particularly challenging. The relative success between species suggests that the acoustic and clustering hyperparameters of our labelling function might need to be tuned for use in other contexts, including other bird species but also possibly mammal, amphibian, or insect sound productions.

*4.2. A modular labelling function*

Our labelling function was designed to be modular. Each part, from segmentation to clustering, is independent and can be adapted to other needs and contexts. Sound segmentation has usually been operated with a single threshold on the time wave (Fodor, 2013; Lasseck, 2013; Potamitis, 2014). Our segmentation method is based on a double threshold process applied to a time-frequency representation of sound. The double threshold process allows the selection of sounds that are not necessarily the most salient ones so that middle-distance sounds can be still detected. Compared to the detector designed by Sprengel et al. (2016), which is widely used to build bird song datasets to train deep learning models such as BirdNET (Kahl et al., 2021), our method segments the signal of interest in time but also in frequency so that the timber can be considered. This strategy opens the possibility of identifying two sounds of interest when they overlap in time but not in frequency. Our segmentation is based on conventional image processing, but deep learning solutions for images have been intensively developed in recent years. These attractive options include temporal convolutional networks (TCNs) (Bai et al., 2018; Steinfath et al., 2021), image object detector architectures such as Faster R-CNN or YOLO (Coffey et al., 2019; Kong et al., 2017; Narasimhan et al., 2017; Shrestha et al., 2021; Venkatesh et al., 2022; Wu et al., 2022; Zsebők et al., 2019) and audio source separation such as speaker diarization (Denton et al., 2022; Morfi and Stowell, 2021). However, segmentation methods all require a very large amount of labelling data as well as high computational resources, two main constraints at a time when naturalist expertise is disappearing and energy consumption must be limited.

To describe the segmented sounds, we opted for hand-crafted features based on the 2D wavelet analysis as proposed by Ulloa et al. (2018). The wavelet transform has the advantage not to be human-centric, contrary to the Mel-frequency cepstral coefficients (MFCC) which has increasing popularity for describing animal vocalizations (Zottesso et al., 2018). In addition, 2D wavelet kernels that are not learned show a striking similarity with convolutional kernels learned by the first-layers of CNN (Meng et al., 2019; Yosinski et al., 2014).

Instead of using hand-crafted features, another approach consists in using learned features, the so-called deep audio embeddings coming from deep learning models already trained on animal vocalizations (Goffinet et al., 2021; Stoumpou et al., 2022). Following this idea, a recent study successfully tested several methods to learn a latent representation of audio samples to separate the song of the barred owl (*Strix varia*), the common crane (*Grus grus*), and the common loon (*Gavia immer*) from noise (Tolkova et al., 2021). The methods include a convolutional autoencoder and two pre-trained networks: the VGGish model (Hershey et al., 2017) and the Wavegram-Logmel-CNN16 architecture (Kong et al., 2020). Both CNN were pre-trained on the Google Audioset dataset, a large collection of annotated sounds from YouTube videos (Gemmeke et al., 2017) with more than 600 classes of sounds, but a unique class "birds". Another solution would have been to use the 512-dimensional embedding provided by BirdNET. In any case, special attention is required to adapt the dimension of the embedding vector to the number of audio samples per species to avoid facing the curse of dimensionality (Assent, 2012). Deep audio embeddings sound promising, but they lack interpretability and still need to be improved by enriching the training dataset with animal signals and soundscapes.

Finally, recent studies proposed using non-linear unsupervised dimensionality reduction such as the Uniform Manifold Approximation and Projection (UMAP) (McInnes et al., 2020) to generate latent space representation directly from the spectrogram. Clustering is then performed directly on the embedding generated with UMAP (Sainburg et al., 2020; Thomas et al., 2021). Although this approach seems very straightforward, it still requires very precise onset detection between calls/songs to avoid intraclass dissimilarity (Thomas et al., 2021).

The last module of the labelling function consisted in separating the sound of interest and noise using DBSCAN clustering. DBSCAN and other density-based clustering non-parametric algorithms have already been used successfully but for different tasks on marine mammals such as separating blue whale (*Balaenoptera musculus*) calls from noise (Cuevas et al., 2017), classifying killer whale (*Orcinus orca*) vocalizations without supervision (Poupard et al., 2019), and identifying the coda repertoire of sperm whales (Amorim et al., 2020; Gero et al., 2016). Very recently, HDBSCAN, a hierarchical variant of DBSCAN (McInnes et al., 2017), has been used to create the vocal repertoire of the meerkat (*Suricata suricatta*) (Thomas et al., 2021), harpy eagles (*Harpia harpyja*), and giant otters (*Pteronura brasiliensis*) (Schneider et al., 2022). So far, only three density-based clustering non-parametric algorithms have been tested (DBSCAN, HDBSCAN and OPTICS [Ankerst et al., 1999]) on animal vocalizations, but many other variants have recently been designed (Latifi-Pakdehi and Daneshpour, 2021) and might be adapted to design more efficient labelling functions.

## 5. Perspectives and conclusion

The growing success of both bioacoustics and ecoacoustics for understanding animal behavior and ecology is mainly based on the collection and automatic analysis of large scale recording archives. If field recording is now made possible with personal recorders and autonomous recording units, the automatic retrieval of sounds of interest for further description is still challenging. Here, taking advantage of the largest bird song audio data set, we showed that a labelling function based on time-frequency analysis and clustering methods can help select appropriate signals and thus improve the quality of the data collected. Further developments are still required to adapt such facilities to soundscape recordings where the sounds of interest of several species are mixed.

**Funding**

This work was realized with the support of the Sorbonne Center for Artificial Intelligence (SCAI) of Sorbonne University (IDEX SUPER 11-IDEX-0004).

**Appendices**

|  | TRAINING DATASET | | | | VALIDATION DATASET | | | | |
|  | DBSCAN | | BirdNET | | | DBSCAN | | BirdNET | |
|  | IN | FN | R | FN | R | IN | FN | R | FN | R |
| --- | --- | --- | --- | --- | --- | --- | --- | --- | --- | --- |
| Long-tailed Tit *(Aegithalos caudatus)* | 55 | 26 | 52 | 38 | 91 | 33 | 0 | 32 | 18 | 96 |
| Eurasian Skylark *(Alauda arvensis)* | 40 | 4 | 71 | 13 | 87 | 48 | 12 | 92 | 12 | 88 |
| Tree Pipit *(Anthus trivialis)* | 24 | 7 | 90 | 5 | 85 | 23 | 14 | 69 | 15 | 94 |
| Short-toed Treecreeper *(Certhia brachydactyla)* | 23 | 2 | 98 | 13 | 100 | 35 | 23 | 92 | 26 | 98 |
| European Greenfinch *(Chloris chloris)* | 55 | 35 | 34 | 31 | 77 | 13 | 4 | 31 | 6 | 75 |
| Common Wood Pigeon *(Columba palumbus)* | 23 | 8 | 28 | 12 | 90 | 69 | 15 | 100 | 29 | 100 |
| Common Cuckoo *(Cuculus canorus)* | 18 | 0 | 12 | 10 | 99 | 25 | 0 | 36 | 9 | 83 |
| Yellowhammer *(Emberiza citrinella)* | 37 | 5 | 47 | 21 | 88 | 47 | 17 | 69 | 21 | 90 |
| European Robin *(Erithacus rubecula)* | 42 | 19 | 67 | 28 | 60 | 36 | 12 | 87 | 13 | 64 |
| Common Chaffinch *(Fringilla coelebs)* | 27 | 2 | 94 | 10 | 88 | 21 | 13 | 80 | 9 | 98 |
| Common Nightingale *(Luscinia megarhynchos)* | 39 | 20 | 90 | 2 | 69 | 40 | 32 | 89 | 22 | 57 |
| Great Tit *(Parus major)* | 54 | 42 | 68 | 25 | 71 | 71 | 38 | 50 | 46 | 100 |
| Common Redstart *(Phoenicurus phoenicurus)* | 71 | 35 | 56 | 57 | 96 | 40 | 18 | 85 | 14 | 90 |
| European Green Woodpecker *(Picus viridis)* | 36 | 4 | 90 | 0 | 76 | 59 | 8 | 71 | 13 | 76 |
| Dunnock *(Prunella modularis)* | 23 | 14 | 83 | 16 | 96 | 26 | 6 | 56 | 14 | 92 |
| Eurasian Nuthatch *(Sitta europaea)* | 39 | 32 | 31 | 31 | 70 | 35 | 24 | 43 | 18 | 82 |
| Tawny Owl *(Strix aluco)* | 15 | 0 | 74 | 6 | 97 | 36 | 0 | 56 | 15 | 91 |
| Eurasian Blackcap *(Sylvia atricapilla)* | 42 | 30 | 100 | 14 | 69 | 42 | 28 | 100 | 23 | 79 |
| Eurasian Wren *(Troglodytes troglodytes)* | 33 | 11 | 77 | 12 | 85 | 27 | 14 | 86 | 10 | 75 |
| Common Blackbird *(Turdus merula)* | 30 | 10 | 57 | 14 | 82 | 31 | 4 | 97 | 4 | 75 |
| Song Thrush *(Turdus philomelos)* | 78 | 76 | 88 | 46 | 54 | 83 | 77 | 95 | 52 | 48 |
| Mistle Thrush *(Turdus viscivorus)* | 27 | 15 | 17 | 16 | 89 | 23 | 10 | 61 | 9 | 80 |

Table A.1. Scores for the train and validation sets for the DBSCAN clustering and BirdNET methods. IN: initial noise, FN: final noise, R: recall.

| | TEST DATASET | | | | |
| --- | --- | --- | --- | --- | --- |
| | | DBSCAN | | BirdNET | |
| | IN | FN | R | FN | R |
| Sedge Warbler *(Acrocephalus schoenobaenus)* | 49 | 27 | 41 | 26 | 52 |
| Eurasian Reed Warbler *(Acrocephalus scirpaceus)* | 67 | 44 | 59 | 37 | 71 |
| Meadow Pipit *(Anthus pratensis)* | 27 | 0 | 37 | 3 | 70 |
| Little Owl *(Athene noctua)* | 58 | 75 | 14 | 38 | 55 |
| European Goldfinch *(Carduelis carduelis)* | 42 | 8 | 70 | 19 | 94 |
| Cetti's Warbler *(Cettia cetti)* | 60 | 45 | 84 | 45 | 84 |
| Eurasian Blue Tit *(Cyanistes caeruleus)* | 80 | 79 | 42 | 68 | 88 |
| Common Reed Bunting *(Emberiza schoeniclus)* | 66 | 64 | 67 | 62 | 83 |
| Barn Swallow *(Hirundo rustica)* | 36 | 20 | 98 | 13 | 94 |
| Northern Wheatear *(Oenanthe oenanthe)* | 23 | 7 | 90 | 9 | 59 |
| Coal Tit *(Periparus ater)* | 37 | 10 | 75 | 23 | 70 |
| Black Redstart *(Phoenicurus ochruros)* | 10 | 3 | 92 | 4 | 85 |
| Common Chiffchaff *(Phylloscopus collybita)* | 47 | 37 | 71 | 36 | 81 |
| Willow Warbler *(Phylloscopus trochilus)* | 29 | 12 | 80 | 13 | 94 |
| Eurasian Bullfinch *(Pyrrhula pyrrhula)* | 82 | 70 | 50 | 20 | 44 |
| Goldcrest *(Regulus regulus)* | 44 | 53 | 25 | 34 | 97 |
| Eurasian Siskin *(Spinus spinus)* | 49 | 41 | 81 | 29 | 84 |
| Eurasian Collared Dove *(Streptopelia decaocto)* | 27 | 5 | 90 | 14 | 95 |
| Common Starling *(Sturnus vulgaris)* | 62 | 45 | 80 | 0 | 14 |
| Garden Warbler *(Sylvia borin)* | 52 | 30 | 95 | 23 | 92 |
| Common Whitethroat *(Sylvia communis)* | 15 | 9 | 84 | 11 | 71 |

| | IN | FN | R | | |
|---|---|---|---|---|---|
| Fieldfare *(Turdus pilaris)* | 57 | 48 | 79 | 28 | 50 |

Table A.2. Scores of the test set for the DBSCAN clustering and the BirdNET methods. IN: initial noise, FN: final noise, R: recall